\documentclass[10pt,twocolumn,letterpaper]{article}

\usepackage[pagenumbers]{cvpr} 

\usepackage{graphicx}
\usepackage{amsmath}
\usepackage{amssymb}
\usepackage{booktabs}
\usepackage{multirow}

\usepackage[pagebackref,breaklinks,colorlinks]{hyperref}

\usepackage[capitalize]{cleveref}
\crefname{section}{Sec.}{Secs.}
\Crefname{section}{Section}{Sections}
\Crefname{table}{Table}{Tables}
\crefname{table}{Tab.}{Tabs.}

\def\cvprPaperID{10078} 
\def\confName{CVPR}
\def\confYear{2022}

\begin{document}

\title{Ranking-Based Siamese Visual Tracking}

\author{Feng Tang$^{1,2}$~ Qiang Ling$^{1}$\thanks{Corresponding Author.}\\
$^1$ Department of Automation, University of Science and Technology of China, China \\
$^2$Institute of Artificial Intelligence, Hefei Comprehensive National Science Center, China\\
{\tt\small tang0420@mail.ustc.edu.cn, qling@ustc.edu.cn}
}

\maketitle

\begin{abstract}
Current Siamese-based trackers mainly formulate the visual tracking into two independent subtasks, including classification and localization. They learn the classification subnetwork by processing each sample separately and neglect the relationship among positive and negative samples. Moreover, such tracking paradigm takes only the classification confidence of proposals for the final prediction, which may yield the misalignment between classification and localization. To resolve these issues, this paper proposes a ranking-based optimization algorithm to explore the relationship among different proposals. To this end, we introduce two ranking losses, including the classification one and the IoU-guided one, as optimization constraints. The classification ranking loss can ensure that positive samples rank higher than hard negative ones, i.e., distractors, so that the trackers can select the foreground samples successfully without being fooled by the distractors. The IoU-guided ranking loss aims to align classification confidence scores with the Intersection over Union(IoU) of the corresponding localization prediction for positive samples, enabling the well-localized prediction to be represented by high classification confidence. Specifically, the proposed two ranking losses are compatible with most Siamese trackers and incur no additional computation for inference. Extensive experiments on seven tracking benchmarks, including OTB100, UAV123, TC128, VOT2016, NFS30, GOT-10k and LaSOT, demonstrate the effectiveness of the proposed ranking-based optimization algorithm. The code and raw results are available at \url{https://github.com/sansanfree/RBO}.

\end{abstract}

\section{Introduction}
\begin{figure}[t]
\centering
\includegraphics[width=\hsize]{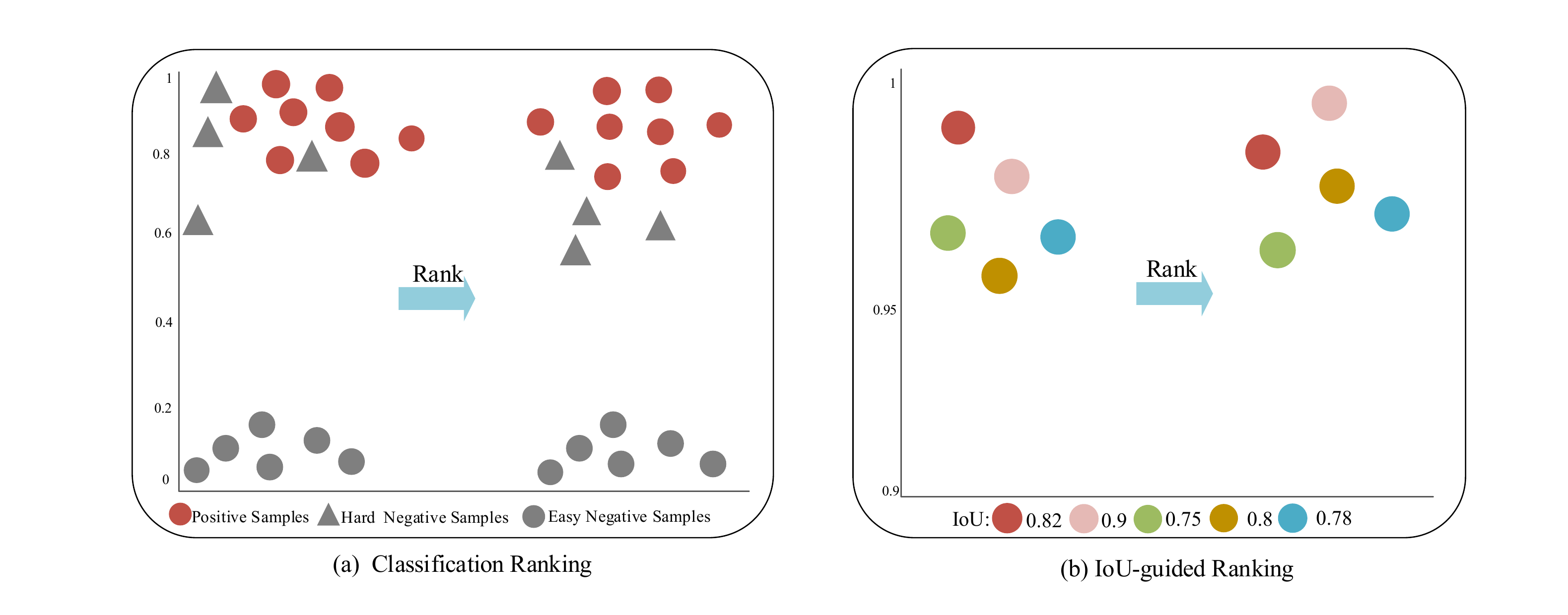}
\caption{Illustration of the proposed two ranking losses. The classification loss enables positive samples to rank higher than hard negative samples, aiming to suppress the classification confidence of distractors. The IoU ranking loss aims to align classification with localization, i.e., that the samples with larger IoUs are expected to gain higher classification confidence scores.}
\label{Fig_1}
\end{figure}
Visual object tracking aims to estimate the location information of an arbitrary target in each frame of a video sequence. In most situations, only the initial target information is provided for trackers, then trackers are required to model the target appearance in the next frames. Since target-specific information is only available at test-time, the target model cannot be obtained via offline network training.

Recently, many researchers exploit how to take the power of deep learning technology to solve the tracking task. Siamese network is one of the most popular deep learning paradigm. As pioneer work, SiamFC\cite{bertinetto2016fully} formulates visual tracking  into a deep target matching problem. To be specific, SiamFC consists of two branches, i.e., target template and search region. Its former branch is used to model the target as a fixed exemplar and the later branch  processes the possible regions. SiamFC inspires many later trackers\cite{wang2018learning,li2018high,li2019siamrpn++,xu2020siamfc++} that are built upon the Siamese network architecture and can achieve state-of-the-art performance. Among them, SiamRPN introduces region proposal networks(RPN) consisting of a classification head for foreground-background discrimination and a regression head for anchor refinement. SiamRPN++\cite{li2019siamrpn++} and SiamFC++\cite{xu2020siamfc++} unleash the
capability of deeper backbone networks, such as ResNet\cite{he2016deep} and GoogleNet\cite{szegedy2015going}, to enhance feature representation. Inspired by anchor-free object detectors like FCOS\cite{tian2019fcos} and CornerNet\cite{law2018cornernet}, many anchor-free trackers\cite{du2020correlation,chen2020siamese,guo2020siamcar,guo2021graph} follow the pixel-wise prediction fashion to perform the target localization.

Although Siamese based trackers have achieved promising performance, there still suffer from two limitations: (1) Siamese trackers have difficulty in distinguishing
background distractors. In particular, at the training stage, the classification subnetwork is optimized by a large number of training samples, among which there exist vast uninformative samples(i.e., easy sample) that can be easily classified while a handful of distracting examples are inundated and contribute to the minor effect on network optimization. At the test-time, although the most non-target samples could be discriminated by the trackers, a background distractor could heavily mislead the tracker when it has strong positive confidence, yielding tracking failure. (2) There exists the mismatch problem between classification and localization as the two tasks are processed separately. More specifically, the classification loss drives the model to distinguish the concerned target from background regardless of the location information, while the regression branch aims at localizing the target's bounding box for all positive samples without taking into account
the classification information. As a result, well-localized proposals may have relatively lower foreground confidence while the proposals with high foreground confidence may yield low localization accuracy.


To resolve the above issues, we propose a ranking-based optimization(RBO) consisting of both classification and IoU-guided ranking losses. The classification loss is used to explicitly model the relationship between the positive and hard negative samples. Actually, there are many diverse sample reweighing strategies to suppress the distractors in object detection\cite{lin2017focal,cai2020learning,cao2020prime}. However, in the context of visual tracking, hard negative samples always have the same semantic class with the tracked target, and it is difficult to discriminate the distractors in the classification embedding space. Differently, as shown in Figure \ref{Fig_1}(a), we tackle the classification as a ranking task where the foreground samples are encouraged to rank higher than background samples. Compared with the original classification loss, the proposed ranking optimization serves a loose constraint, under which hard negative samples are allowed to be classified as the foreground as long as their foreground confidence scores are lower than those of positive ones, and can well prevent the tracker from being fooled by distractors. Actually, the misalignment problem between classification and localization was studied in object detection\cite{liu2021rankdetnet,zhang2021varifocalnet,wang2021reconcile}, inspired by which we propose an IoU-guided ranking loss on the basis of RankDetNet\cite{liu2021rankdetnet} to align the foreground confidence scores of with their corresponding IoU values as shown by Figure \ref{Fig_1}(b). The modified loss is more suitable for tracking task, which enables classification confidence to be localization-sensitive to some extent.

To verify the effectiveness of the proposed ranking-based optimization, we choose anchor-based SiamRPN++\cite{li2019siamrpn++} and anchor-free SiamBAN\cite{chen2020siamese} trackers as baselines, and craft our SiamRPN++-RBO and SiamBAN-RBO trackers, respectively. Moreover, recent works propose diverse pixel-wise correlation ways to calculate the similarity between the exemplar and search images\cite{liao2020pg,guo2021graph,fu2021stmtrack}. Inspired by them, we modify the SiamBAN-RBO by replacing the depth-wise correlation with the pixel-wise correlation, obtaining a new tracker version called SiamPW-RBO.
The main contributions of this paper are summarized as follows.
\begin{itemize}
\item We devise a classification ranking loss to enhance the discrimination capability by modeling the relationship between foreground samples and background ones, which prevents the tracker from being fooled by distractors.
\item We propose an IoU-guided ranking loss to alleviate the mismatch problem between classification and localization. It connects two independent subtasks by aligning the classification score with associated IoU, ensuring that well-localized prediction could be represented by high classification confidence score.
 \item The proposed RBO can significantly improve the performance of three types of trackers on seven diverse benchmarks without sacrificing inference speed compared with baseline trackers.
\end{itemize}

\section{Related Work}\label{sec:RW}
\begin{figure*}[t]
\centering
\includegraphics[width=\hsize]{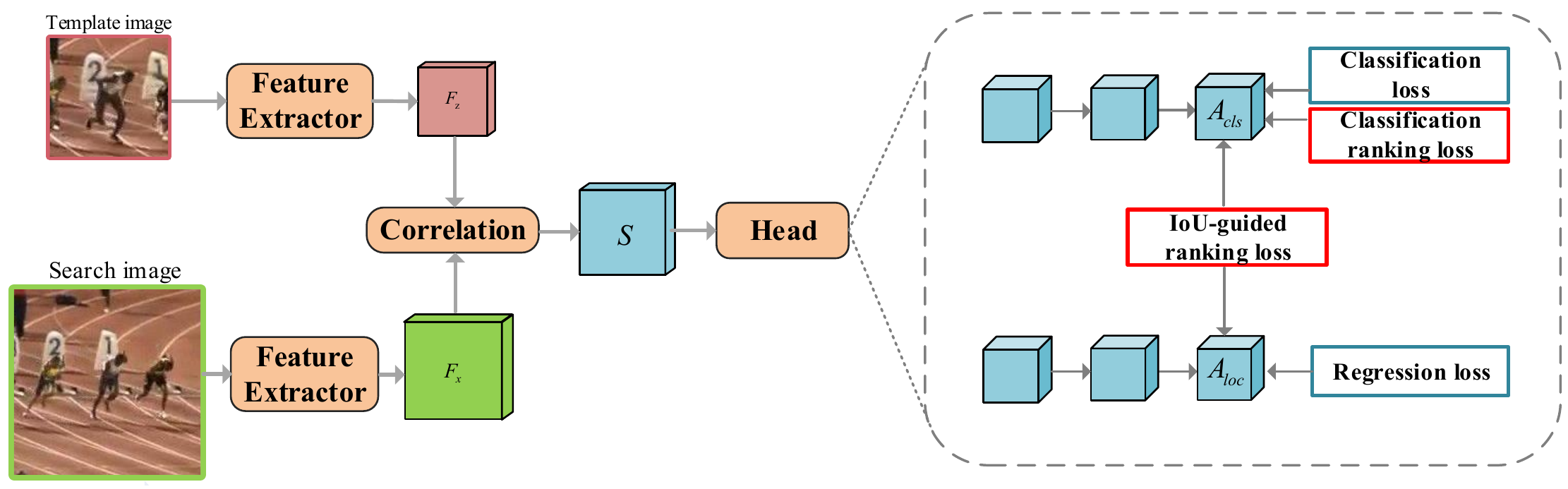}
\caption{The pipeline of the Siamese network based tracker consists of two subtasks, including classification and localization. The proposed classification ranking loss  facilitates the classification optimization while the IoU-guided ranking loss aims to align classification confidence with localization prediction.}
\label{Fig_2}
\end{figure*}

\subsection{Siamese visual tracking}
Recently, SiamFC\cite{bertinetto2016fully} formulates the visual tracking task as a general similarity computation problem between the target template and the search region, which learns a general discriminator via large-scale offline training. Subsequent trackers have been proposed to further enhance the Siamese framework by introducing an attention mechanism\cite{wang2018learning}, designing a new network architecture\cite{wang2019spm}, using an enhanced loss\cite{dong2018triplet}, or utilizing deep reinforcement learning\cite{wang2020post}. In these follow-ups, it is worth mentioning that SiamRPN\cite{li2018high} introduces the RPN module to predict bounding boxes of the targets undergoing variation in aspect ratio, instead of the brute-force discrete scale search strategy of SiamFC. Hence, SiamRPN upgrades SiamFC into an advanced framework. Based on SiamRPN, various trackers were proposed. Among them, DaSiamRPN\cite{zhu2018distractor} collects more diverse training data to enhance discrimination. C-RPN\cite{fan2019siamese} constructs multi-stage RPNs to perform state estimation more accurately. SiamRPN++\cite{li2019siamrpn++} adopts a deeper ResNet-50\cite{he2016deep} network to enhance feature representation. Inspired by the anchor-free object detection and instance segmentation, some trackers modified the original RPN architecture into pixel-wise tracking\cite{xu2020siamfc++,chen2020siamese,guo2020siamcar,guo2021graph}. Besides that, \cite{yu2020deformable,liao2020pg,guo2021graph,fu2021stmtrack} focus on facilitating similarity learning via modification on non-local attention\cite{wang2018non}.

Although the above Siamese trackers have achieved satisfactory performance, they can be easily misled by distractors, i.e., the robustness of the trackers can be weak. To resolve this robustness issue, many researchers introduce online deep learning techniques to enhance the generalization capability of trackers. For example, ATOM\cite{danelljan2019atom} and DiMP\cite{bhat2019learning} construct a target-special classifier at each online tracking, and collect the historical hard negative samples to boost the classifier. UpdateNet\cite{zhang2019learning} updates the target template to incorporate the temporal information. MAML\cite{wang2020tracking} transfers object detectors, such as FCOS\cite{tian2019fcos}, to trackers via meta-learning and incrementally updates the network weights to adapt the target-special sequences. However, those trackers need to design online tracking protocols carefully to avoid false and redundant updates. Differently, this paper proposes a rank-based classification loss to suppress distractors and only involves offline training without modification on network architectures. Therefore, the proposed rank loss is computation-free for inference.
\subsection{Misalignment Problem in Visual Tracking}
Most of advanced trackers process classification and localization separately, ignoring the misalignment problem between them. Some researchers adopt feature alignment strategies to perform scale-aware correlation. For example, Ocean\cite{zhang2020ocean} introduces deformable convolution\cite{dai2017deformable} and SAM\cite{zhang2018visual} utilizes spatial transformer networks.
Besides them, many trackers perform re-detection to achieve more accurate localization\cite{wang2019spm,voigtlaender2020siam}. However, they usually utilize pool operation and design complicated re-detection mechanism, resulting in high complexity. SiamRCR\cite{peng2021siamrcr} and SiamLTR\cite{tang2021learning} adds an additional branch to evaluate the localization prediction, aiming to achieving localization-sensitive proposal selection criterion.
Different from them, we propose an IoU-ranking loss to facilitate back-propagation, which aims to align the confidence score with the associated IoU. It is worth mentioning that our tracker does not need to add any extra network architecture or design any new tracking protocol. So it is totally cost-free at the inference stage.

\subsection{Ranking Algorithms}
Learning to rank has been widely used in NLP tasks such as recommendation systems, and aims to optimize the ranking of sample lists. Recently, some researchers consider ranking optimization in visual object detection. For example, AP-loss\cite{chen2019towards} is proposed to optimize the average precision metric of classification directly. DR loss\cite{qian2020dr} abandons original classification loss and optimizes the ranking of foreground and background distributions. RankDetNet\cite{liu2021rankdetnet} employs two ranking constraints for classification and localization, respectively.
However, to our knowledge, learning to rank has not been implemented to visual tracking. Although our ranking-based optimization shares partial similarity with the above methods, the motivation and technical details are quite different: (1) The optimization tasks are different. Ranking strategies in object detection aim to learn class information for the task-special object detection task while our ranking optimization is tailored to enhance similarity measurement utilized in the class-agnostic tracking task. (2) The implementations are different. Our ranking optimization serves as an additional constraint on the original loss, instead of replacing the original ones. (3) The purposes of ranking are different. Ranking strategies utilized in object detection expect that the inter-class variance is large while the intra-class variance is small, which means the samples which belong to the same class should be similar in the classification space. On the contrary, our classification ranking loss is used to enlarge the intra-class distance as the target and hard distractors always have the same class.

\section{Method}\label{sec:PM}
In this section, we will introduce the proposed ranking-based optimization(RBO) based on the Siamese network based trackers. Figure \ref{Fig_2} shows the overall pipeline.
Firstly, we will briefly review the Siamese trackers in Section \ref{sec:rst}. Then, we will present the proposed RBO in the following sections.
\subsection{Revisiting Siamese Trackers}\label{sec:rst}
The standard Siamese trackers take an exemplar image $z$ and a search image $x$ as input. The image $z$ points out the concerned target in the first frame, and the trackers are required to locate the target in the search region $x$ in subsequent video frames. The two images are fed into a shared backbone network to generate feature maps $\boldsymbol{F}_{z} \in \mathbb{R}^{H_{z} \times W_{z} \times C}$ and $\boldsymbol{F}_{x} \in \mathbb{R}^{H_{x} \times W_{x} \times C}$, respectively.  Then a matching
network $\varphi$ is applied to process $\boldsymbol{F}_{z}$ and $\boldsymbol{F}_{x}$ to obtain the similarity feature map $\boldsymbol{S}$ as
\begin{equation} \label{Eq_1}
\boldsymbol{S}=\varphi\left(\boldsymbol{F}_{z}, \boldsymbol{F}_{x}\right)
\end{equation}
Many popular Siamese trackers define $\varphi$ as depth-wise cross-correlation(DW-Corr)\cite{dong2020clnet,yu2020deformable,li2019siamrpn++,xu2020siamfc++,chen2020siamese,guo2020siamcar}. Recently, inspired by the video object segmentation\cite{oh2019video}, many researchers take pixel-wise correlation methods(PW-Corr)\cite{liao2020pg,guo2021graph,fu2021stmtrack}, which are variants of non-local attention\cite{wang2018non}, as the matching network $\varphi$ of the tracking task. In this paper, we introduce a simplified version of PW-Corr\cite{fu2021stmtrack} as
\begin{equation} \label{Eq_2}
w_{i j}=\frac{\exp \left[\left(\boldsymbol{F}_{z}^{i} \odot \boldsymbol{F}_{x}^{j}\right) / \sqrt{C}\right]}{\sum_{\forall k} \exp \left[\left(\boldsymbol{F}_{z}^{k} \odot \boldsymbol{F}_{x}^{j}\right) / \sqrt{C}\right]}
\end{equation}
where $\boldsymbol{F}_{z}$ and $\boldsymbol{F}_{x}$ are reshaped into the size of $H_{z}W_{z} \times C$ and $C\times H_{x}W_{x}$ , respectively, and $i$ and $j$ are the indices of each pixel on $\boldsymbol{F}_{z}$ and $\boldsymbol{F}_{x}$, respectively. The symbol $\odot$ denotes the dot-product operation. Then we obtain a similarity matrix $w \in \mathbb{R}^{H_{z}W_{z} \times H_{x}W_{x}}$. The similarity feature map $\boldsymbol{S}$ is calculated as
\begin{equation} \label{Eq_3}
\boldsymbol{S}=\operatorname{concat}\left(\boldsymbol{F}_{x},\left(\boldsymbol{F}^{z}\right)^{T} \otimes w\right)
\end{equation}
where $\operatorname{concat}()$ represents matrix concatenation, and $\otimes$ denotes matrix multiplication. Then, the similarity feature map $\boldsymbol{S}$ is fed into the RPN head which consists of classification module $\theta_{cls}$ and localization module $\theta_{loc}$. The RPN head could be anchor-based\cite{li2018high,li2019siamrpn++} or anchor-free\cite{chen2020siamese,du2020correlation}. We can obtain the classification map $\boldsymbol{A}_{cls}$ and the regression map $\boldsymbol{A}_{loc}$  as
\begin{equation}\label{Eq_4}
\boldsymbol{A}_{cls}=\mathcal{\theta}_{cls}\left(\boldsymbol{S}\right), \quad \boldsymbol{A}_{loc}=\mathcal{\theta}_{loc}\left(\boldsymbol{S}\right)
\end{equation}
$\boldsymbol{A}_{cls}$ aims to identify the foreground proposals from background ones while $\boldsymbol{A}_{reg}$ regresses the target bounding box. The standard loss of Siamese trackers is defined as
\begin{equation}\label{Eq_5}
\begin{aligned}
\mathcal{L}_{rpn}&=\frac{1}{N_{{pos}}} \sum_{i \in \mathcal{A}_{pos}}\mathcal{L}_{cls}\left(\boldsymbol{A}_{cls}^{i}, \boldsymbol{Y}_{cls}^{i}\right)+\mathcal{L}_{loc}\left(\boldsymbol{A}_{loc}^{i}, \boldsymbol{Y}_{loc}^{i}\right)\\
&+\frac{1}{N_{{neg}}} \sum _{i \in \mathcal{A}_{neg}} \mathcal{L}_{{cls}}\left(\boldsymbol{A}_{cls}^{i}, \boldsymbol{Y}_{cls}^{i}\right)
\end{aligned}
\end{equation}
where $N_{pos}$ and $N_{neg}$ are the numbers of the positive sample set $\mathcal{A}_{pos}$ and the negative sample set $\mathcal{A}_{neg}$, respectively. $\boldsymbol{Y}_{cls}$ and $\boldsymbol{Y}_{loc}$ denote the classification and regression labels, respectively. $\mathcal{L}_{cls}$ is usually the cross entropy loss and $\mathcal{L}_{loc}$ is the commonly used smooth $L_1$ loss or IoU loss.

We can observe two limitations from Eq. \ref{Eq_5}. Firstly, the classification branch processes the positive and negative samples separately and does not explore the relationship between them. Secondly, the classification and localization branches are trained with independent objective functions without direct interactions between them, which may yield prediction inconsistency between the classification and localization.

\subsection{Classification Ranking Loss}\label{sec:cr}
\begin{figure}[t]
\centering
\includegraphics[width=\hsize]{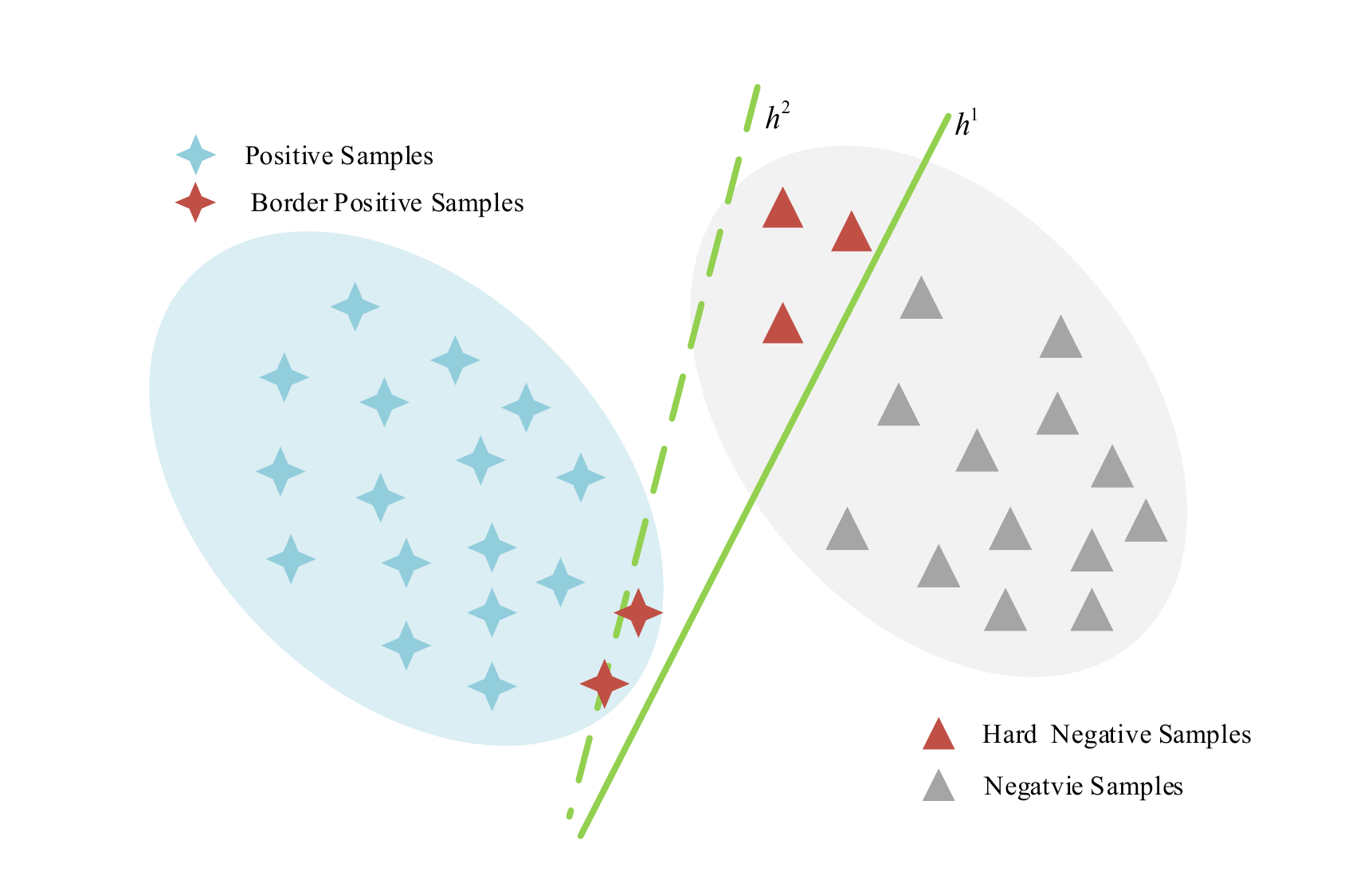}
\caption{Illustration of binary classification without or with the proposed classification ranking loss. $h^1$(solid green line) and $h^2$(dotted green line) are the corresponding decision hyperplanes.}
\label{Fig_3}
\end{figure}
As discussed above, most Siamese-based trackers achieve binary classification via cross entropy loss which can ensure that most samples could be classified correctly. However, as shown in Figure \ref{Fig_3}, some hard negative samples may cross the decision hyperplane and fool the classifier. In the tracking task, as long as the classification score of one negative sample is larger than those of all positive samples, tracking failure occurs. Hence, false positive classification severely hampers the robustness of trackers.

\begin{figure}[t]
\includegraphics[width=\hsize]{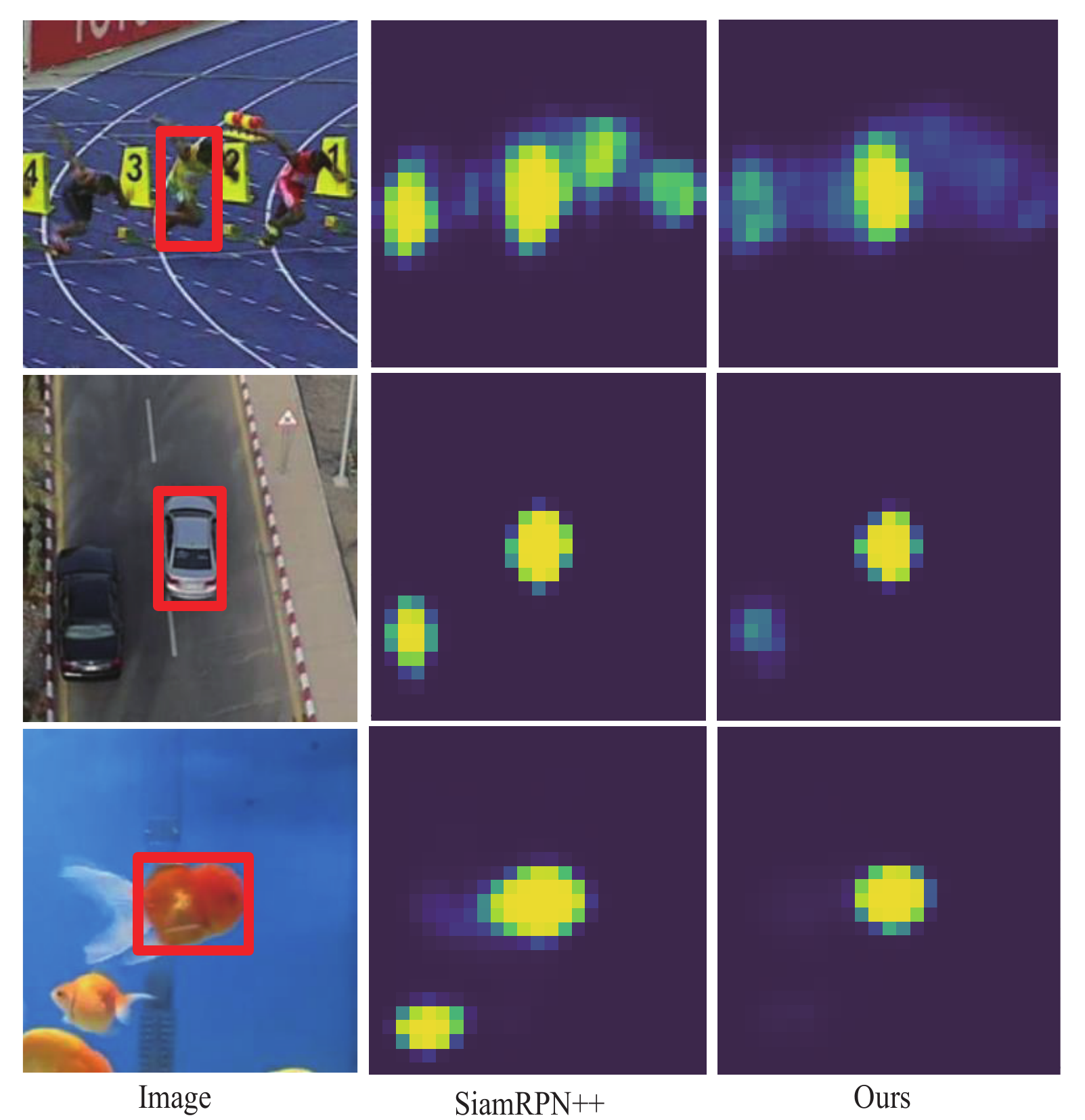}
\caption{ Confidence maps of the target object (red box) estimated by SiamRPN++ and Ours(SiamRPN++ supervised by the proposed classification ranking loss). The model predicted by SiamRPN++ struggles to distinguish the target from distractor objects in the background. In contrast, the proposed ranking loss provides superior discriminative power for SiamRPN++, and can suppress the confidence of distractors significantly.}
\label{Fig_4}
\end{figure}

To alleviate this issue, we propose a classification ranking loss to enlarge the foreground-background classification decision margin. In particular, we first train the classifier, which is supervised by cross entropy loss. Then we sort all the negative samples by their predicted object confidence scores. The negative samples whose confidence scores are lower than $\tau_{neg}$, e.g., 0.5, are filtered out. The rest ones constitute the hard negative sample set $\{p_{j_-}\}_{j_-}^{n_-}$, where $n_-$ is the number of negative samples, $p_{j_-}$ denotes the object confidence score of sample $j_-$. Analogously, for the positive samples, we keep all of them to obtain the positive set $\{p_{j_+}\}_{j_+}^{n_+}$. As for the next ranking optimization, we do not employ a point-wise comparison between the set $\{p_{j_-}\}_{j_-}^{n_-}$ and $\{p_{j_+}\}_{j_+}^{n_+}$ due to two reasons. Firstly, the time complexity of the point-wise comparison is equal to $\mathcal{O}(n_-n_+)$, which is expensive for training. Secondly, it is not necessary that each positive sample should rank higher than all negative samples, as some low-confidence positive samples, which are located at the classification borderlines, could be ignored to some extent. Moreover, as long as one positive sample ranks higher than the hard negative ones, the tracker can select the right candidate as the tracking target. In light of the above considerations, we rank the expectations of training samples to enlarge the foreground-background classification margin while the time complexity can be significantly reduced to $\mathcal{O}(1)$. The expectation of hard negative and positive samples are defined as
\begin{eqnarray}\label{Eq_6}
\begin{split}
P_{-}=\sum_{j_-}^{n_-} w_{j_-} p_{j_-} \\
P_{+}=\sum_{j_+}^{n_+} w_{j_+} p_{j_+}
\end{split}
\end{eqnarray}
where $w_{j_-}$ denotes the expectation weight of sample $j_-$, $w_{j_-}$ is normalized by the SoftMax function as
\begin{equation}\label{Eq_7}
w_{j_-}=\frac{\exp ( p_{j_-} )}{\sum_{j_-}^{n_-} \exp (p_{j_-})}
\end{equation}
Differently, the positive weight $w_{j_+}$ is set as $\frac{1}{n_+}$ since we want to preserve the positive classification distribution.
We adopt the logistic loss to rank the expectations $P_{-}$ and $P_{+}$ as
\begin{equation} \label{Eq_8}
\mathcal{L}_{\text{rank-{cls}}}=\frac{1}{\beta} \log (1+\exp (\beta \cdot (P_{-}-P_{+}+\alpha))
\end{equation}
where $\beta$ controls the loss value, and $\alpha$ is a ranking margin. Specifically, if there are no hard negative samples in an image, we will skip this image. As shown in Figure \ref{Fig_3}, with the supervision of $\mathcal{L}_{\text {rank-{cls}}}$, the decision hyperplane is adjusted from $h^1$ to $h^2$ and the hard negative samples are placed at the negative side successfully. Note that some border positive samples may be located at the negative side of the decision hyperplane, which is acceptable for the single object tracking task as we just need one positive sample to represent the tracked target. In addition, Figure \ref{Fig_4} further visualizes the effectiveness of the proposed classification ranking loss, where high responses only reflect on the concerned target while the distractors are suppressed significantly.

\subsection{IoU-Guided Ranking Loss}\label{sec:igr}
To tackle the mismatch between classification confidence and localization, we propose an IoU-guided ranking loss to harmonize the optimization of classification and regression branches. More specifically, the proposed loss aims to align the confidence scores of two positive samples with their associated IoU, and can encourage one positive sample with a larger IoU to rank higher than the other one with a smaller IoU. To this end, for the positive sample ${i,j \in \mathcal{A}_{pos}}$, the ranking constraints are organized in a pair-wise manner as
\begin{eqnarray}\label{Eq_9}
\begin{split}
  p_i>p_j, \quad  s.t.\; v_{i}^{iou}>v_{j}^{iou} \\
  v_{i}^{iou}>v_{j}^{iou}, \quad  s.t.\; p_i>p_j
\end{split}
\end{eqnarray}
where $p_i$ and $p_j$ indicate the foreground confidence scores of positive samples $i ,j$, respectively, $ v_{i}^{iou}$ and $ v_{j}^{iou}$ denote the predicted IoU values with the ground-truth of samples $i$ and $j$. Note that our ranking constraints are modified from RankDetNet\cite{liu2021rankdetnet}, and their difference will be analyzed in Section \ref{sec:as}. Then the IoU-guided ranking loss is defined as
\begin{equation}\label{Eq_10}
\begin{aligned}
L_{\text{rank-iou}}&=\frac{1}{N_{pos}}{\sum \limits_{{i,j \in \mathcal{A}_{pos}}, v_{i}^{iou}> v_{j}^{iou}} exp(-\gamma \cdot (p_i-p_j)) } \\
&+\frac{1}{N_{pos}}{\sum \limits_{{i,j \in \mathcal{A}_{pos}}, p_{i}> p_{j}} exp(-\gamma \cdot (v_{i}^{iou}- v_{j}^{iou})) }
\end{aligned}
\end{equation}
where $\gamma>0$ is a hyper-parameter to control the loss value. During the back-propagation optimization process, if $v_{i}^{iou}> v_{j}^{iou}$, we will optimize $p_i$ and $p_j$ to make $p_i$ to rank higher than $p_j$; if $p_i> p_j$, following \cite{liu2021rankdetnet}, we will freeze $v_{j}^{iou}$ and only optimize $v_{i}^{iou}$ to achieve the expected ranking. If $v_{j}^{iou}$ is not frozen, the loss could drop by decreasing $v_{j}^{iou}$, which would hamper regression optimization.

The proposed IoU-guided ranking loss can narrow the gap between the classification and regression branches by aligning classification scores with the associated IoUs. Thereby well-localized prediction could be represented by high classification confidence.
\subsection{Ranking-Based Trackers}\label{sec:rbt}
Advanced Siamese trackers employ different backbone networks, correlation ways and RPN heads. Since our proposed ranking-based optimization(RBO) aims to facilitate classification and regression optimization, RBO is not sensitive to network architectures. For convenience, we adopt ResNet-50\cite{he2016deep} as backbone network, and integrate the proposed RBO into SiamRPN++(DW-Corr and anchor-based head)\cite{li2019siamrpn++} and SiamBAN(DW-Corr and anchor-free head)\cite{chen2020siamese}, obtaining SiamRPN++-RBO and SiamBAN-RBO, respectively. Moreover, we craft a new version named as SiamPW-RBO by replacing DW-Corr with the introduced PW-Corr in Section \ref{sec:rst} on the basis of SiamBAN-RBO.

As shown in Figure \ref{Fig_2}, the proposed two ranking losses can be optimized together with the original loss adopted in Siamese trackers. We empirically combine the original loss $\mathcal{L}_{\text {RPN}}$ with the proposed $\mathcal{L}_{\text{rank-{cls}}}$ and $\mathcal{L}_{\text{rank-{iou}}}$ with the weights of 1:0.5:0.25, which benefits a stable offline training. Since the proposed RBO only involves offline training, it does not introduce any extra computation cost at the inference stage.

\section{Experiments}\label{sec:ex}
Our trackers are implemented using the Pytorch tracking platform PySOT and trained on four NVIDIA GTX 1080Ti GPUs.

\textbf{Implementation Details}. For a fair comparison, we follow the same training protocols(datasets and training hyper-parameters) defined in PySOT for SiamRPN++-RBO, SiamBAN-RBO and SiamPW-RBO. For both training and inference, template patches are resized to the size of 127$\times$127 pixels and the search regions are cropped to 255$\times$255 pixels. The ranking margin $\alpha$ is set as 0.5 according to the analysis of \cite{qian2020dr}. In order to achieve a stable training, $\beta$ and $\gamma$ are set as 4 and 3, respectively, for all experiments.

\textbf{Evaluation Datasets and Metrics.}
We use seven tracking benchmarks, including OTB100\cite{wu2015object}, UAV123\cite{mueller2016benchmark}, NFS30\cite{kiani2017need}, TC128\cite{liang2015encoding},  GOT-10k\cite{huang2019got}, VOT2016\cite{kristan2016visual} and LaSOT\cite{fan2019lasot} for the tracking performance evaluation. For VOT2016\cite{kristan2016visual}, we adopt the accuracy(A), robustness(R) and Expected Average Overlap(EAO) metrics. For GOT-10k\cite{huang2019got}, the trackers are evaluated on its online server, which employs the average overlap (AO) and success rate (SR) metrics. For other datasets, we adopt distance precision(DP) at the 20 pixels and area-under-curve (AUC) score of overlap success plots for evaluation.
\subsection{Comparison with State-of-the-art Trackers}
\begin{table*}
\resizebox{\textwidth}{!}{
\begin{tabular}{cccccccccccccc}
\hline
&SiamRPN++&SiamBAN&SiamCAR &Ocean&CLNet&CGACD&SiamRN & SiamRPN++ &SiamBAN&SiamGAT &\textbf{SiamRPN++}&\textbf{SiamBAN}&\textbf{SiamPW}\\
& \cite{li2019siamrpn++}&\cite{chen2020siamese}&\cite{guo2020siamcar}&\cite{zhang2020ocean}&\cite{dong2020clnet}&\cite{du2020correlation} & \cite{cheng2021learning} & -ACM\cite{han2021learning} &-ACM\cite{han2021learning}& \cite{guo2021graph}&\textbf{-RBO}&\textbf{-RBO}&\textbf{-RBO}\\
\hline
OTB100\cite{wu2015object}&69.6&69.6&65.7&67.2& 65.7& \color[rgb]{0, 0, 1}71.3 &70.1& \color[rgb]{0, 1, 0} 71.2 &\color[rgb]{1, 0, 0}72.0& 71.0& 69.9&  70.1 &69.8  \\
TC128\cite{liang2015encoding} &57.3&58.4 &57.8 & 55.1& 56.4 &\color[rgb]{0, 1, 0}60.5&-& -&-&58.5&\color[rgb]{1, 0, 0}61.9  & \color[rgb]{0, 0, 1}61.2&59.3\\
UAV123\cite{mueller2016benchmark}&61.3&61.4 &61.4 &59.2& 63.3& 63.3&\color[rgb]{0, 1, 0}64.3&  63.4&\color[rgb]{1, 0, 0}64.6& \color[rgb]{0, 0, 1}64.5& \color[rgb]{0, 1, 0}64.3& 64.1&\color[rgb]{0, 0, 1} 64.5 \\
NFS30\cite{kiani2017need} &50.2&59.0 &53.3 & 51.8& 54.3 &55.4&-& -&-&56.7&\color[rgb]{0, 1, 0}59.6  & \color[rgb]{1, 0, 0}61.3&\color[rgb]{0, 0, 1}60.1\\

\hline
\end{tabular}}
\caption{Comparison results on on the OTB100\cite{wu2015object}, TC128\cite{liang2015encoding}, UAV123\cite{mueller2016benchmark} and NFS30\cite{kiani2017need} datasets in terms of AUC score. Red, blue and green indicate top three results. Ocean is the offline version. }
\label{table_four}
\end{table*}

\begin{table*}
\resizebox{\textwidth}{!}{
\begin{tabular}{ccccccccccccc}
\hline
&SiamRPN++&Ocean&D3s&SiamFC++&SiamBAN &SiamCAR&KYS & STMTrack &SiamGAT &\textbf{SiamRPN++}&\textbf{SiamBAN}&\textbf{SiamPW}\\
&\cite{li2019siamrpn++}&\cite{zhang2020ocean}&\cite{lukezic2020d3s}&\cite{xu2020siamfc++}& \cite{chen2020siamese}&\cite{guo2020siamcar}&\cite{bhat2020know} & \cite{fu2021stmtrack}  & \cite{guo2021graph}&\textbf{-RBO}&\textbf{-RBO}&\textbf{-RBO}\\
\hline
AO($\uparrow$) &51.7 & 59.2& 59.7 &59.5&57.9& 56.9&\color[rgb]{0, 1, 0}63.6&\color[rgb]{0, 0, 1}64.2  & 62.7&60.2&60.8&\color[rgb]{1, 0, 0}64.4\\
SR$_{0.5}$($\uparrow$)&61.5&69.5 &67.6 &69.5& 68.4 & 67.0&\color[rgb]{0, 0, 1}75.1&  73.7&  \color[rgb]{0, 1, 0}74.3& 71.8&  72.2 &\color[rgb]{1, 0, 0} 76.7 \\
SR$_{0.75}$($\uparrow$)&32.9&47.9&46.2&47.3& 45.7& 41.5 &\color[rgb]{0, 0, 1}51.5&  \color[rgb]{1, 0, 0}57.9 & 48.8& 44.6&  46.8 &\color[rgb]{0, 1, 0}50.9  \\
\hline
\end{tabular}}
\caption{Comparison results on the GOT-10k\cite{huang2019got} test set in terms of average overlap (AO), and success rates (SR) at the overlap thresholds of 0.5 and 0.75.}
\label{table_got10k}
\end{table*}

\textbf{OTB100\cite{wu2015object}}. We validate our proposed trackers on the OTB100 dataset\cite{wu2015object}, which consists of 100 fully annotated sequences. As shown in Table \ref{table_four}, our proposed SiamRPN++-RBO, SiamBAN-RBO and SiamPW-RBO achieve the AUC scores of 69.9$\%$, 70.1$\%$ and 69.8$\%$, respectively. Compared with the recent proposed Siamese trackers such as SiamRN\cite{cheng2021learning} and SiamGAT\cite{guo2021graph}, our three trackers achieve better or competitive performance against them.

\textbf{TC128\cite{liang2015encoding}}. For further evaluation, we report tracking results on the TC128 dataset consisting of 128 color sequences. As shown in Table \ref{table_four}, the SiamBAN-RBO and SiamRPN++-RBO outperform existing state-of-the-art Siamese trackers such as SiamGAT\cite{guo2021graph}. Moreover, SiamRPN++-RBO  significantly boosts the AUC score of the baseline SiamRPN++ from 57.3$\%$ to 61.9$\%$.

\textbf{UAV123\cite{mueller2016benchmark}}. UAV123 dataset contains 123 low-altitude aerial videos captured from a UAV. This dataset has numerous sequences
with partial or full occlusions and drastic deformation. As shown in Table \ref{table_four}, our SiamRPN++-RBO obtains a success (AUC) score of 0.643, which significantly outperforms the baseline SiamRPN++ with a large margin. The significant improvement also happens to the SiamBAN-RBO version. This is because the RBO method can increase the difference between representations of the target and background, which contributes to distinguishing the target from distractors.

\textbf{NFS30\cite{kiani2017need}}. We conduct experiments on NFS dataset(30 FPS version)\cite{kiani2017need}, which provides 100 challenging videos with fast-moving objects. As shown in Table \ref{table_four}, our three trackers steadily rank top three and outperform recent Siamese trackers like SiamGAT\cite{guo2021graph}.

\textbf{GOT-10k\cite{huang2019got}}. The recently released GOT-10k dataset provides a large-scale and high-diversity benchmark where the \emph{training} and \emph{testing} subsets have no overlapping. Following its protocol, we just use its \emph{training} subset to train our models and do evaluation on its \emph{testing} set. Table \ref{table_got10k} shows the results in terms of average overlap (AO) and success rate (SR) metrics with the overlap thresholds of 0.5 and 0.75. SiamPW-RBO slightly outperforms other top-performing trackers such as memory-based STMTrack\cite{fu2021stmtrack} in term of AO metric. Compared with SiamRPN++, our SiamRPN++-RBO obtains significant improvements of 8.5$\%$ in AO, 10.3$\%$  in SR$_{0.5}$ as well as 11.7$\%$ in SR$_{0.75}$. The improvement is partially attributed to the fact that the proposed IoU-guided ranking loss can enable trackers to estimate the target states more accurately and alleviate the issue of error accumulation.

\textbf{VOT2016\cite{kristan2016visual}}. We compare our trackers on VOT2016 in Table \ref{table_vot16}. VOT2016 contains 60 challenging sequences and ranks the trackers by EAO. As Table \ref{table_vot16} reports, our proposed SiamRPN++-RBO outperforms the baseline SiamRPN++ with an absolute gain of 4.9$\%$ in EAO.  Analogously, SiamBAN-RBO achieves favorable performance against SiamBAN with an EAO of 0.543.
\begin{table}[t] \small
\centering
\begin{tabular}{l|cccc}
\hline
Tracker &A($\uparrow$) &R($\downarrow$)  &EAO($\uparrow$) \\ \hline
\hline
SiamRPN\cite{li2018high}&0.578& 0.312&0.337\\
Da-SiamRPN\cite{zhu2018distractor}&0.596& 0.266&0.364\\
SiamDW\cite{zhang2019deeper}&0.580& 0.240&0.371\\
SiamMask-\emph{Opt}\cite{wang2019fast}&\color[rgb]{0, 0, 1}0.670& 0.230&0.442\\
UpdateNet\cite{zhang2019learning}&0.610& 0.206&0.481\\
SiamRPN++\cite{li2019siamrpn++}&0.642&0.196& 0.463\\
Ocean\cite{zhang2020ocean}&0.625& 0.158&0.486\\
ROAM++\cite{yang2020roam}&0.559&0.174 &0.441  \\
SiamBAN\cite{chen2020siamese}&0.632&0.149&0.502\\
D3s\cite{lukezic2020d3s}&0.667&0.158&0.499\\
SiamRPN++-ACM\cite{han2021learning}&\color[rgb]{0, 1, 0}0.666&0.144&0.501\\
SiamBAN-ACM\cite{han2021learning}&0.647&\color[rgb]{1, 0, 0}0.098&\color[rgb]{1, 0, 0}0.549\\
SiamAttn\cite{yu2020deformable}&\color[rgb]{1, 0, 0}0.680&\color[rgb]{0, 1, 0}0.140&\color[rgb]{0, 1, 0}0.537\\\hline
SiamRPN++-RBO\textbf{(Ours)}&0.635&\color[rgb]{0, 1, 0}0.140 &0.512\\
SiamBAN-RBO\textbf{(Ours)}&0.629&\color[rgb]{0, 0, 1}0.112&\color[rgb]{0, 0, 1}0.543 \\
SiamPW-RBO\textbf{(Ours)}&0.617&\color[rgb]{1, 0, 0}0.098&0.531 \\ \hline
\end{tabular}
\caption{Comparison with state-of-the-art trackers on the VOT2016 dataset\cite{kristan2016visual} in terms of accuracy(A), robustness(R) and expect average overlap(EAO).}
\label{table_vot16}
\end{table}

\begin{figure}[t]
\centering
\includegraphics[width=\hsize]{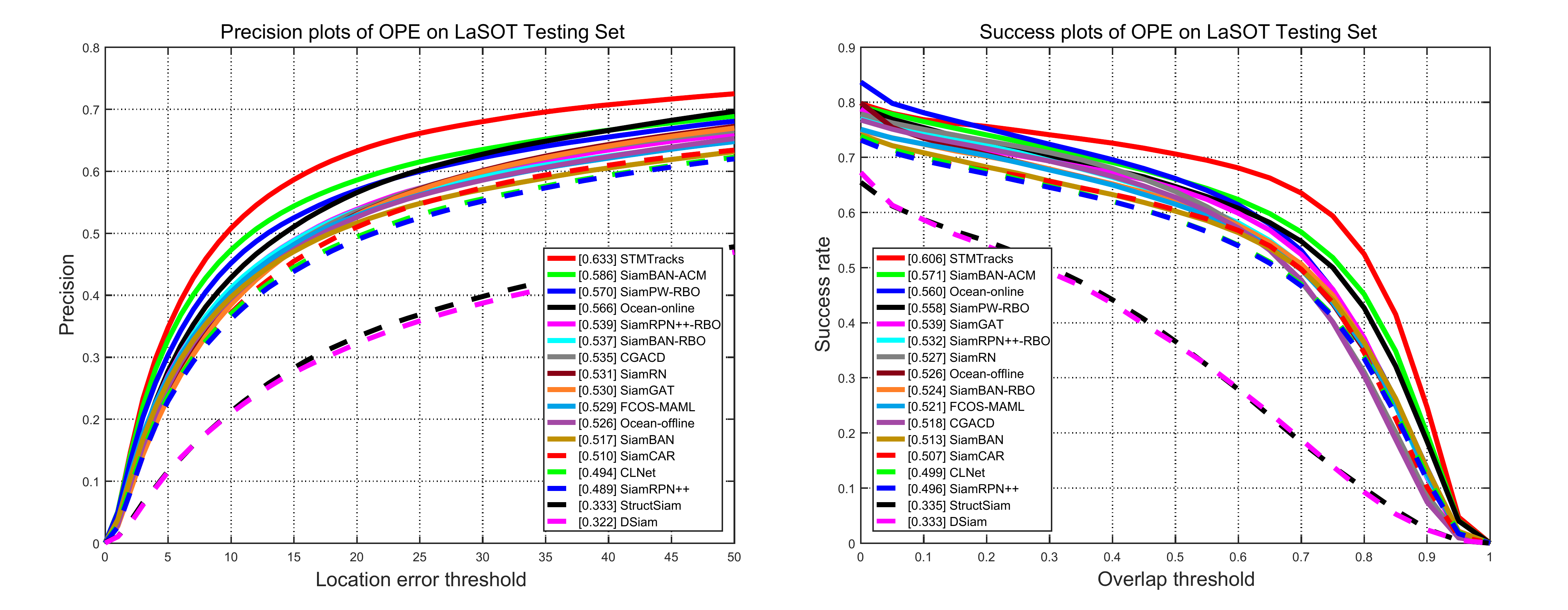}
\caption{Precision and success plots on the LaSOT\cite{fan2019lasot} \emph{testing} dataset.}
\label{Fig_5}
\end{figure}

\textbf{LaSOT\cite{fan2019lasot}}. LaSOT is also a large-scale, high-quality dataset. Its \emph{testing} set contains 280 test sequences with an average length of 2506 frames, and is often used to evaluate the long-term tracking performance. As Figure \ref{Fig_5} depicts, Our SiamRPN++-RBO and SiamBAN-RBO improve the AUC scores of baselines with the gains of 3.6$\%$ and 1.1$\%$, respectively.

\subsection{Ablation Study}\label{sec:as}
In this section, we perform extensive analysis of the proposed three versions on the combined dataset containing the entire TC128\cite{liang2015encoding}, NFS30\cite{kiani2017need} and UAV123\cite{mueller2016benchmark} datasets. This pooled dataset contains 352 diverse sequences to enable thorough analysis. The AUC score is adopted for evaluation.

\textbf{Ranking-based Optimization(RBO).}
We perform a thorough ablation study on each component of RBO. As Table \ref{table_4} shows, classification ranking loss(CR) improve the AUC scores of SiamRPN++ and SiamBAN by 2.70$\%$ and 1.54$\%$, respectively. The significant gains demonstrate that CR, which models the relationship between the positive samples and hard negative samples, could help to learn a more discriminative classifier. Furthermore, when the trackers are equipped with the IoU-guided ranking loss(IGR), IGR brings the gains of 1.45$\%$(from 60.63$\%$ to 62.08$\%$) and 1.09$\%$ from (61.15$\%$ to 62.24$\%$) for the SiamRPN++ and SiamBAN versions, respectively.

\begin{figure}[t]
\centering
\includegraphics[width=\hsize]{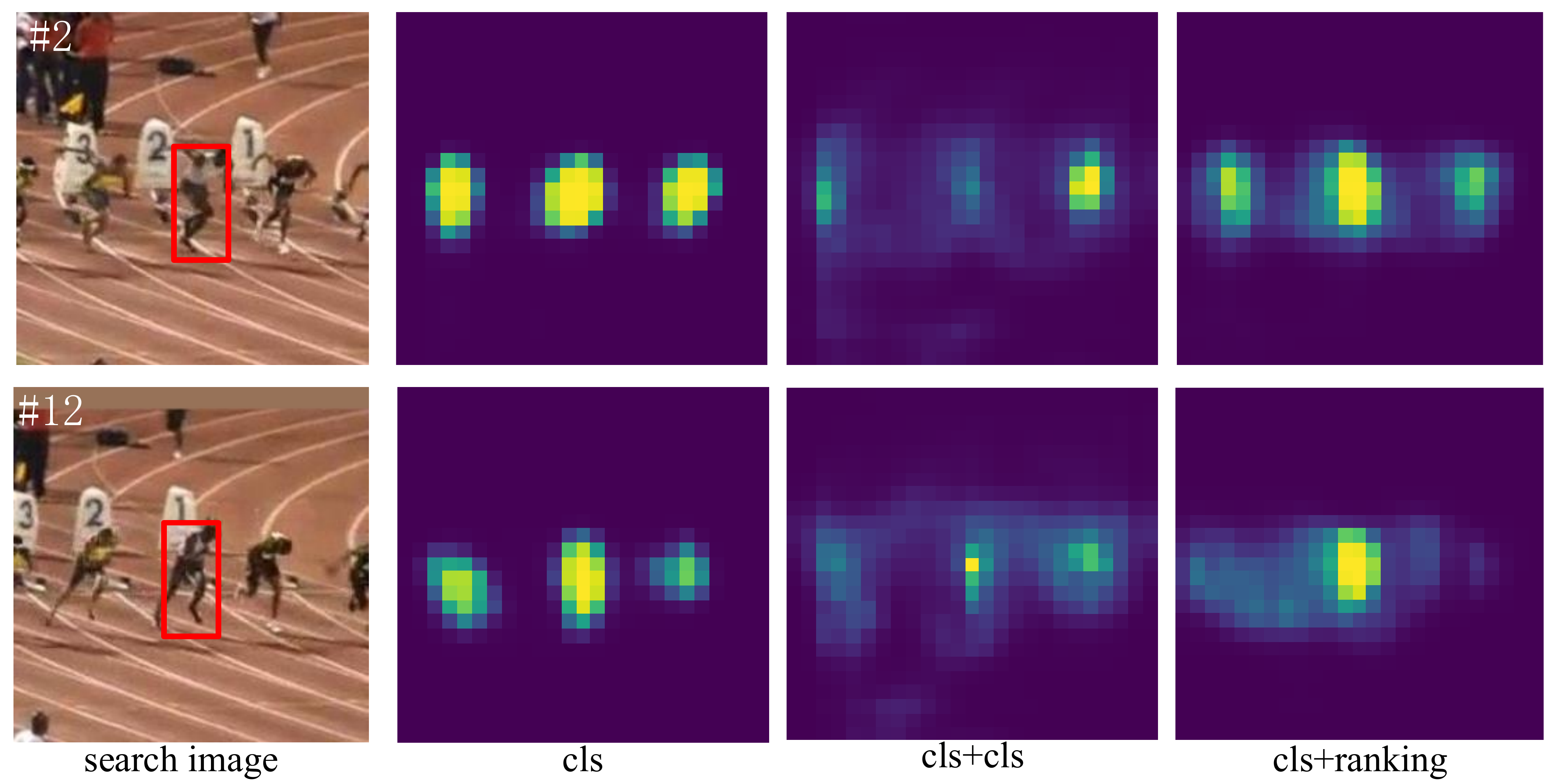}
\caption{Visualization of confidence maps. From left to right: search image, confidence maps with single-stage classification loss, two-stage classification loss and
the classification loss+the proposed ranking loss, respectively.}
\label{Fig_6}
\end{figure}

\textbf{The Strategy of Hard Negative Mining.}
We test a regular hard negative mining strategy, i.e., employing two-stage cross entropy losses, among which the first one selects hard negative samples and the second one aims to optimize these hard samples. As shown in Figure \ref{Fig_6}, unfortunately, the real concerned target is suppressed along with the non-target regions, and the classifier still struggles to distinguish the target from the distractors. On the contrary, the proposed method, i.e., cross entropy loss+ranking loss, not only highlights the concerned target, but also suppresses the hard distractors, which confirms that our CR is effective in improving the discrimination of the classifier.
\begin{table}
\centering
\resizebox{\columnwidth}{!}{
\begin{tabular}{ccc|cc}
\hline
CR&IGR-ori &IGR&SiamRPN++\cite{li2019siamrpn++}&SiamBAN\cite{chen2020siamese}\\ \hline
 &&&57.93&59.61 \\
 $\checkmark$ &&&60.63 &61.15 \\
$\checkmark$&$\checkmark$ &&60.87 &60.92\\
$\checkmark$& &$\checkmark$&62.08& 62.24 \\
 \hline
\end{tabular}}
\caption{Ablation analysis of the proposed ranking-based optimization, consisting of classification ranking(CR), original IoU-guided ranking\cite{liu2021rankdetnet} (IGR-ori)  and our IoU-guided ranking(IGR) losses on the combined TC128, NFS30 and UAV123 datasets.}
\label{table_4}
\end{table}

\textbf{Difference with RankDetNet\cite{liu2021rankdetnet}.}  As shown in Table \ref{table_4}, the original IoU-guided ranking loss in \cite{liu2021rankdetnet}, i.e., $\mathcal{L}=\mathcal{L}_{\text {rank }}(-\alpha \cdot\left(p_{i}-p_{j}\right)(v_{i}^{iou}-v_{j}^{iou}))$, fails to boost our method. We argue that it is not easy for $\mathcal{L}$ to optimize the four variables($p_i$, $p_j$, $v_{i}^{iou}$, $v_{j}^{iou}$) together since the relationship among the four variables is missing and they may be updated along sub-optimal directions. Differently, in our modified loss(Eq. \ref{Eq_10}), the joint optimization is divided into two subtasks, and there are only two variables to optimize under the explicit constraint(Eq. \ref{Eq_9}) in each iteration. As Table \ref{table_4} shows, our modified loss can further lift the performance, which confirms that more strong and explicit supervision may be more suitable for class-agnostic visual tracking training.
\subsection{Evaluation on the Transformer trackers}
To further evaluate the RBO on the Transformer trackers\cite{chen2021transformer,wang2021transformer,yan2021learning,yu2021high}, we choose the TransT\cite{chen2021transformer} method for comparison. From Table \ref{table_5}, TransT+RBO version outperforms the TransT on the three datasets. This indicates that although the Transformer can model the relationship of different proposals by attention mechanism, the RBO can still provide additional cues for facilitating offline optimization.

\begin{table}
\centering
\resizebox{\columnwidth}{!}{
\begin{tabular}{|c|ccc|ccc|cc|}
\hline
\multirow{2}{*}{Tracker} &\multicolumn{3}{c|}{LaSOT\cite{fan2019lasot}}&\multicolumn{3}{c|}{GOT-10k\cite{huang2019got}}&\multicolumn{2}{c|}{UAV123\cite{mueller2016benchmark}}\\
\cline{2-9}
&AUC &$P_{Norm}$&P&AO &SR$_{0.5}$&SR$_{0.75}$&DP&AUC \\\hline
TransT&64.9&73.8&69.0 &72.3&82.4&68.2&87.4&67.9\\
 TransT+RBO &65.6&74.3 &69.7&72.7&82.9&68.7&88.0&68.5 \\
\hline
\end{tabular}}
\caption{Comparison with the TransT\cite{chen2021transformer} and TransT+RBO methods on the LaSOT, GOT-10k and UAV123 datasets.}
\label{table_5}
\end{table}
\section{Conclusion and Discussion}
In this paper, we propose a ranking-based optimization algorithm for Siamese tracking. Firstly, we propose a classification ranking loss, which converts the classification optimization into ranking problems where the positive samples are encouraged to rank higher than hard negative ones. After the involvement of ranking optimization, the tracker can select the top-rank positive sample as the concerned target without being fooled by distractors. Moreover, in order to reconcile consistency of prediction between classification and localization, we propose an IoU-ranking loss to optimize the classification and localization tasks together at the offline training stage, thereby producing the target estimation with co-occurrence of high classification confidence and localization accuracy at the inference.

\textbf{Limitations.} From Table \ref{table_5}, we observe that the performance gain of our RBO on Transformer tracker is inferior to that on Siamese trackers\cite{li2019siamrpn++,chen2020siamese}. In the future, we intend to explore more advanced ranking strategies to boost the Transformer based methods further.

\textbf{Acknowledgements} This work was supported in part by Technological Innovation Project of the New Energy and Intelligent Networked Automobile Industry of Anhui Province(Research, Development and Industrialization of Intelligent Cameras), and in part by the Key Science and Technology Project of Anhui Province under Grant 202203f07020002.

{\small
\bibliographystyle{ieee_fullname}
\bibliography{10078}
}

\end{document}